\documentclass{article}
\usepackage{spconf,amsmath,epsfig}
\usepackage{cite}
\usepackage{multirow}
\usepackage{multicol}

\title{GUIDED DEEP LEARNING BY SUBAPERTURE DECOMPOSITION: OCEAN PATTERNS FROM SAR IMAGERY}

\name{Nicolae-C\u{a}t\u{a}lin Ristea$^1$, Andrei Anghel$^1$, Mihai Datcu$^{1,2}$, Bertrand Chapron$^3$ \thanks{This work was supported by IFREMER and by the grant of the Romanian Ministry of Education and Research, CNCS - UEFISCDI, project number PN-III-P4-ID-PCE-2020-2120, within PNCDI III.}}

\address{CEOSpaceTech, University Politehnica of Bucharest, Romania$^1$ \\ Remote Sensing Technology Institute, German Aerospace Center (DLR), Germany$^2$ \\ Laboratoire d’Ocanographie Physique et Spatiale (LOPS), Ifremer, Brest, France$^3$}

\begin{document}

\maketitle

\begin{abstract}
Spaceborne synthetic aperture radar (SAR) can provide meters-scale images of the ocean surface roughness day-or-night in nearly all weather conditions. This makes it a unique asset for many geophysical applications. Sentinel-1 SAR wave mode (WV) vignettes have made possible to capture many important oceanic and atmospheric phenomena since 2014. However, considering the amount of data
provided, expanding applications requires a strategy to automatically process and extract geophysical parameters. In this study, we propose to apply subaperture decomposition (SD) as a preprocessing stage for SAR deep learning models. Our data-centring approach surpassed the baseline by 0.7\%, obtaining state-of-the-art on the TenGeoP‐SARwv data set. In addition, we empirically showed that SD could bring additional information over the original vignette, by rising the number of clusters for an unsupervised segmentation method. Overall, we encourage the development of data-centring approaches, showing that, data preprocessing could bring significant performance improvements over existing deep learning models.
\end{abstract}


\section{Introduction}
\label{sec:intro}
\vspace{-0.2cm}

The ocean covers more than 70\% of the Earth's surface, conditioning fundamentally the climate system. Comprehensive observations and measurements of the ocean surface are essential in order to have a better understanding of air–sea interactions as well as to develop high‐resolution climate models \cite{Topouzelis-2015-RSE}. One of the most used space-borne sensors for ocean observation is the SAR, used by satellite mission Sentinel-1 from 2014, when the WV mode, dedicated for retrieving ocean wave proprieties at global scale \cite{Stopa-2016-GRL}, was implemented. The routine WV measurements, available only on the Sentinel-1A/B, have a spatial resolution of approximately 4 meters and a scene footprint of 20 by 20 km. These sensors collect monthly nearly 120,000 WV vignettes of the global ocean surface, but without automated means to identify the geophysical features captured by each image, the potential would remain untapped.

Several works have been proposed for automatic analysis of vignettes in order to extract interpretative information, which could be transformed in geophysical parameters. On the one hand, classic machine learning algorithms have mostly been developed for oil spills and ships detection \cite{Almulihi-2021-RS}. These methods depend on the empirically hand-crafted features, which are usually insufficient to generalize the local variations, shapes and structural patterns \cite{Topouzelis-2015-RSE, Zhang-2016-GRSM}. On the other hand, once with the development of deep neural networks (DNNs), deep learning algorithms have become more popular in geo-science related tasks, such as, ocean SAR imagery \cite{Wang-2019-RSE, Quach-2020-TRGS, Colin-2021-IGARSS, De-2021-JSTAR}. In \cite{Quach-2020-TRGS} authors proposed an automatic system for significant wave height prediction from SAR vignettes, which surpassed the existing methods by a significant margin. They developed a DNN architecture to fusion late features from two separate branches. Another example of DNN for ocean examination is a fully convolutional network (FCN) used to predict the sea ice concentration \cite{De-2021-JSTAR}. The network is based on a U-Net architecture and is able to obtain an accuracy of 78.2\%, while classifying between 6 classes.
In order to extract global information from vignettes, in \cite{Wang-2019-RSE} authors propose to classify each ocean surface vignette in accordance with 10 geophysical phenomena, annotated in TenGeoP‐SARwv data set \cite{Wang-2019-GDJ}. They obtained an overall accuracy of 98.4\% by fine-tuning the InceptionV3 network on decimated intensity vignettes. Different from them, we propose to use the SD algorithm as a preprocessing stage in order to enhance the network's performance.

Nevertheless, TenGeoP‐SARwv data set contains a single label for observations which covers an oceanic region of 20 square km. This categorization might be adequate for large scale physical phenomena (e.g., wind streaks), but it is not applicable for local phenomena (e.g., icebergs). Moreover, the shapes of the ocean features are also diverse, like narrow curves for fronts, aggregation of disconnected areas for biological slicks or wide regions for low wind areas \cite{Colin-2021-IGARSS}. To overpass this rough labeling generalization, in \cite{Colin-2021-IGARSS} authors propose a semantic segmentation task, conducting to regions from the same vignette which presents the same geophysical phenomena. They rely on a FCN architecture based on U-Net, which is able to predict, at pixel level, the classes from the input vignettes. Differently, we adapted an unsupervised segmentation algorithm \cite{Kim-2020-TIP}, which is able to find multiple clusters in the same vignette. Not only that the number of clusters for our approach is not upper bounded by the number of classes presented in the data set, like in \cite{Colin-2021-IGARSS}, but also we do not need any labels to perform the segmentation. Moreover, we empirically showed that by decomposing vignettes in subapertures, the model is able to find more clusters, enforcing the idea that observing the ocean from different angles, we might see different backscatter patterns.

The SD algorithm is widely used for SAR imagery \cite{Wang-2020-GRSL, Brekke-2013-GRSL, Singh-2010-IGARSS, Focsa-2020-COMM}. The method was combined with both classical signal processing algorithms \cite{Brekke-2013-GRSL, Singh-2010-IGARSS, Focsa-2020-COMM} and deep learning methods \cite{Wang-2020-GRSL}. In \cite{Brekke-2013-GRSL} the SD algorithm is proposed for ship detection, while in \cite{Singh-2010-IGARSS} it is used for target characterization. Moreover, the SD algorithm was used to transform a single channel SAR image into three channels image, by decomposing into three subapertures, in order to use pretrained DNN for target classification on the ground \cite{Wang-2020-GRSL}. Distinct for all SD based approaches, we propose to use this algorithm as a generic preprocessing step for training SAR deep learning methods on ocean vignettes.


In summary, our contribution is twofold:
\begin{itemize}
\vspace{-0.2cm}
\item We are the first who propose the SD algorithm as an preprocessing stage for ocean SAR deep learning models, achieving state-of-the-art results on TenGeoP‐SARwv data set.
\vspace{-0.2cm}
\item We adapted an unsupervised segmentation method \cite{Kim-2020-TIP} for ocean SAR imagery and empirically showed that training models on subapertures, rather than the original vignettes, enrich the number of classes found on the ocean surface.
\end{itemize}
\vspace{-0.5cm}

\section{Method}
\vspace{-0.2cm}

\begin{figure*}[!t]
\begin{center}
\centerline{\includegraphics[width=0.9\linewidth]{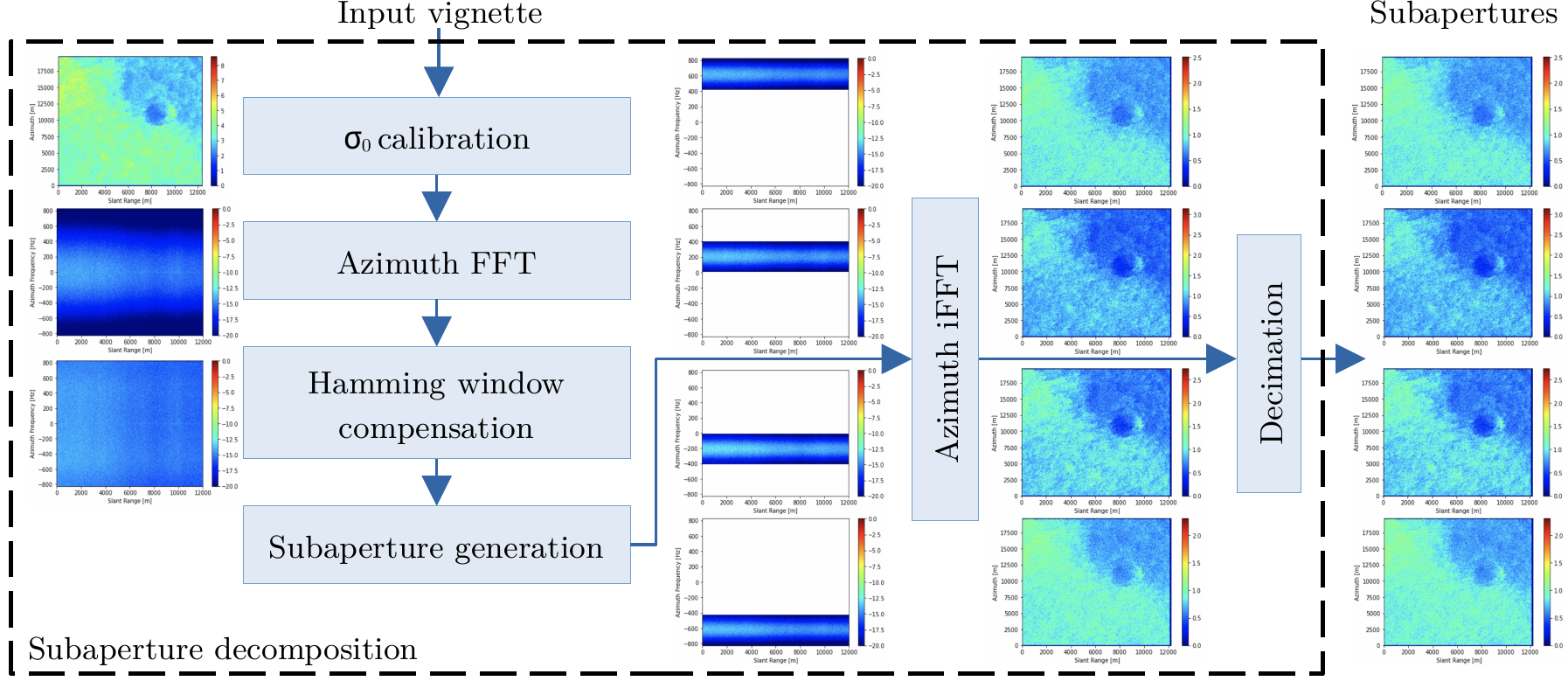}}
\vskip -0.3cm
\caption{The preprocessing subaperture decomposition pipeline. An input vignette is processed by a series of blocks, followed by the subaperture generation. Next, each subaperture is processed by the azimuth inverse FFT and the decimation blocks. The output of each block is presented in an associated image.}
\label{fig_pipeline}
\end{center}
\vskip -1.0cm
\end{figure*}

\subsection{Subaperture decomposition}

The basic SAR system acquires the backscatter returned from irradiated targets in different positions and different azimuth angles along the radar trajectory. In order to obtain high azimuth resolution, the real antenna aperture is replaced by the synthetic aperture, by processing the signal along the azimuth dimension. Considering that the ocean surface is highly non-stationary, observing it from different angles might bring additional information about the illuminated area.
Consequently, we define the subaperture as the image formed using only a part of the total azimuth angle. Therefore, decomposing the original vignette into multiple subapertures, we can mimic different observation angles of the same scene. The SD algorithm is visually described in Fig. \ref{fig_pipeline}.

\noindent
\textbf{The $\boldsymbol{\sigma_0}$ calibration.}
According to \cite{Wang-2019-GDJ}, the measured normalized radar cross section $\sigma_0$ by SAR over the ocean is highly dependent on the local ocean surface wind and viewing angles (incidence and azimuth) of the radar. Therefore, the $\sigma_0$ of each input vignette is calibrated by dividing it to a reference factor, constructed by assuming a constant wind of $10$ m/s at $45^{\circ}$ relative to the antenna look angle.

\noindent
\textbf{Azimuth FFT.}
After $\sigma_0$ calibration, we perform the Fast Fourier Transform (FFT) along the azimuth axis, in order to obtain the vignette's spectrum. The number of FFT points is equal to the number of points in the azimuth direction.

\noindent
\textbf{Hamming window compensation.}
Next, we perform a Hamming window compensation, with a coefficient of $0.75$, in order to obtain a flat azimuth spectrum.

\noindent
\textbf{Subaperture generation.}
In the following stage, we filter the processed vignette with $4$ shifted Hamming windows (with the same $0.75$ coefficient), in order to obtain the corresponding azimuth spectrum for each subaperture.

\noindent
\textbf{Azimuth iFFT.}
Having the azimuth spectrum for each subaperture, we want to translate back the data into time domain by performing an inverse Fast Fourier Transform (iFFT), with the same parameters from the \textit{Azimuth FFT} block.

\noindent
\textbf{Decimation.}
The last stage of the SD algorithm is the decimation. The fine‐resolution SAR subapertures are not necessary for large scale geophysical phenomena, especially since the classes described in \cite{Wang-2019-GDJ} have scales of tens to thousands of metres. Therefore to better highlight larger feature patterns, for each resulted subaperture, we compute the intensity image followed by a low-pass-filtering with a window of $10\times10$, each filter's coefficient being $0.01$. The intensity images are then decimated by $1/10$ yielding a resolution of 50 meters.
\vspace{-0.5cm}

\subsection{Deep learning tasks}
\vspace{-0.2cm}

\noindent
\textbf{Classification.}
The success of the convolutional neural networks (CNNs) in image processing tasks \cite{Dhillon-2020-PAI} encouraged their introduction in remote sensing applications and SAR imagery \cite{Wang-2019-RSE, Quach-2020-TRGS, Colin-2021-IGARSS, De-2021-JSTAR}. Proposing a data-centring approach, we focused our attention on the preprocessing stage, rather than the network's architecture, therefore we employed two well-known architectures, ResNet18 and InceptionV3, for the ocean SAR image classification task. The networks were pretrained on the ImageNet data set and only two architectural changes were made: the number of output neurons (10) and the number of input channels (in accordance with the input).

\noindent
\textbf{Unsupervised segmentation.}
We employed the solution presented in \cite{Kim-2020-TIP} for unsupervised ocean SAR images segmentation. Considering an input SAR image, the pixel labels and feature representations are jointly optimized, by updating the parameters with the gradient descent algorithm. Pixel label prediction and network's parameters learning are alternately iterated to meet the following three conditions: (a) pixels of similar features should be assigned the same label, (b) spatially continuous pixels should be assigned the same label and (c) the number of unique labels should be large. To serve our scope, we adapted the first convolutional layer of the network in accordance with the input's number of  channels.

\vspace{-0.3cm}
\section{Experiments}
\vspace{-0.2cm}

\noindent
\textbf{Data set.}
TenGeoP‐SARwv data set contains over 37,000 ocean vignettes with 10 geophysical phenomena. We used the raw vignettes from the TenGeoP‐SARwv data set, with the assigned labels, and randomly split the data in training (70\%), validation (15\%) and test (15\%).

\begin{table}
\centering
\caption{Accuracy and inference time results on the TenGeoP‐SARwv test set. The input type symbols are: O for original vignette and S$_i$ for the i$^{th}$ subaperture. 
The significantly better results (level $0.01$) than corresponding baselines, according to a paired McNemar's test, are marked with $\dagger$.}

\label{tab_results}
 \begin{tabular}{|l|cccc|}
 \hline
  Method & Input & Accuracy & \multicolumn{2}{c|}{Time (ms)}\\
         & & &                           CPU & GPU \\
 \hline\hline
 InceptionV3 \cite{Wang-2019-RSE} & O                & 98.4  & 101       &   12    \\ 
 \hline
 InceptionV3 & S$_1$            & 94.8  & 121       &   18    \\
 \hline
 InceptionV3 & S$_{1,2,3,4}$      & 99.1$^{\dagger}$  & 163       &   21    \\
 \hline
 InceptionV3 & O + S$_{1,2,3,4}$ & 99.1   & 165       &   21    \\

 \hline
 \hline

 ResNet18 & O                   & 98.0      & 41        & 2     \\ 
 \hline
 ResNet18 & S$_1$               & 94.0      & 56        & 5     \\
 \hline
 ResNet18 & S$_{1,2,3,4}$         & 98.9$^{\dagger}$    & 98        & 12     \\
 \hline
 ResNet18 & O + S$_{1,2,3,4}$     & 98.9      & 101       & 12      \\
 \hline

\end{tabular}
\end{table}

\noindent
\textbf{Hyper-parameters tuning.}
We tuned the hyper-parameters on the validation set. The number of subapertures was considered from 2 to 6, the best results being achieved for 4. For the classification task, we trained the models for 30 epochs with Adam optimizer and a mini-batch size of 32 samples. We set the learning rate to $10^{-4}$ and we used a weight decay of $10^{-5}$. For \cite{Kim-2020-TIP}, we used the parameters proposed in the paper.

\noindent
\textbf{Results.}
In Table \ref{tab_results} we present the results obtained for two DNN models, ResNet18 and InceptionV3, on TenGeoP‐SARwv test set. By using only one subaperture, corespondent to a quarter of the entire spectrum, both models have a drop of $4\%$ in accuracy. When all 4 subapertures are used (S$_{1,2,3,4}$), the accuracy raise with $0.7\%$ for InceptionV3 and $0.8\%$ for ResNet18, in comparison with the corresponding baselines. But, when the subapertures are concatenated with the original vignette, no other improvements are observed. In addition, we compared the DNNs inference time (including SD time) in order to observe the overhead brought by our preprocessing method. The SD algorithm is slower with approximately 60 ms on a Intel i9 CPU and with only 10 ms on a nVidia RTX3090 card. But, the $0.5\%$ performance improvement brought by ResNet18 model trained on subapertures, in comparison with the baseline from \cite{Wang-2019-GDJ}, comes with no processing time drawbacks.

\begin{figure}
\begin{center}
\centerline{\includegraphics[width=1\linewidth]{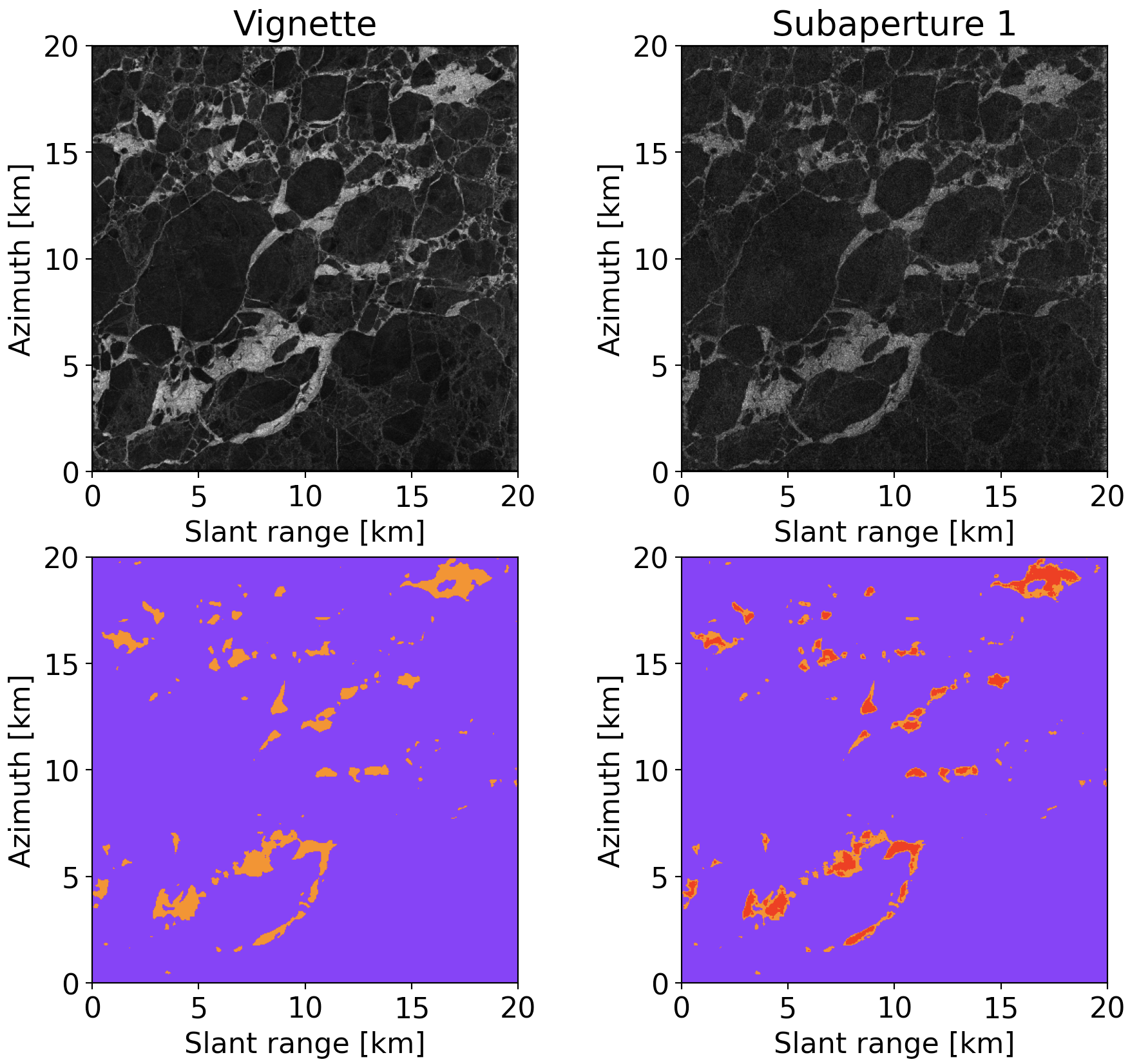}}
\vskip -0.2cm
\caption{Segmentation results for \cite{Kim-2020-TIP} when the input is the original vignette (left) and the concatenated subapertures (right). The number of classes found are: 2 (left) and 3 (right).}
\label{fig_segmentation}
\end{center}
\vskip -1cm
\end{figure}

Moreover, in Fig.~\ref{fig_segmentation} we showed qualitative results for the unsupervised segmentation task on the ocean surface. In the left column, is presented the vignette with the associated segmentation, where only 2 different classes have been found. While, in the right column, is presented the subaperture 1 with the segmentation result (for the segmentation we considered at input all subapertures), containing 3 different classes. Consequently, using SD preprocessing method, deep models may find new classes on the ocean SAR data, helping to locate and characterise more geophysical phenomena.

\vspace{-0.2cm}
\section{Conclusion}
\vspace{-0.2cm}

In this paper, we proposed the SD algorithm as a data-centring preprocessing method for ocean SAR images, which can improve DNNs performances in two related tasks. We showed that better results could be achieved using shallower nets, with no additional processing time, and the SD could help segmentation methods to find more classes on the ocean surface. In future works, we aim to analyse more SAR processing algorithms, which can be used for DNNs. 

\vspace{-0.25cm}

\bibliographystyle{IEEEbib}
\bibliography{main}

\begin{thebibliography}{10}

\bibitem{Topouzelis-2015-RSE}
K.~Topouzelis and D.~Kitsiou,
\newblock ``Detection and classification of mesoscale atmospheric phenomena
  above sea in sar imagery,''
\newblock {\em Remote Sensing of Environment}, vol. 160, pp. 263--272, 2015.

\bibitem{Stopa-2016-GRL}
J.~E. Stopa, F.~Ardhuin, R.~Husson, H.~Jiang, B.~Chapron, and F.~Collard,
\newblock ``Swell dissipation from 10 years of envisat advanced synthetic
  aperture radar in wave mode,''
\newblock {\em Geophysical Research Letters}, vol. 43, no. 7, pp. 3423--3430,
  2016.

\bibitem{Almulihi-2021-RS}
A.~Almulihi, F.~Alharithi, S.~Bourouis, R.~Alroobaea, Y.~Pawar, and
  N.~Bouguila,
\newblock ``Oil spill detection in sar images using online extended variational
  learning of dirichlet process mixtures of gamma distributions,''
\newblock {\em Remote Sensing}, vol. 13, no. 15, pp. 2991, 2021.

\bibitem{Zhang-2016-GRSM}
L.~Zhang, L.~Zhang, and B.~Du,
\newblock ``Deep learning for remote sensing data: A technical tutorial on the
  state of the art,''
\newblock {\em IEEE Geoscience and Remote Sensing Magazine}, vol. 4, no. 2, pp.
  22--40, 2016.

\bibitem{Wang-2019-RSE}
C.~Wang, P.~Tandeo, A.~Mouche, J.~E. Stopa, V.~Gressani, N.~Longepe,
  D.~Vandemark, R.~C Foster, and B.~Chapron,
\newblock ``Classification of the global sentinel-1 sar vignettes for ocean
  surface process studies,''
\newblock {\em Remote Sensing of Environment}, vol. 234, pp. 111457, 2019.

\bibitem{Quach-2020-TRGS}
B.~Quach, Y.~Glaser, J.~E. Stopa, A.~A. Mouche, and P.~Sadowski,
\newblock ``Deep learning for predicting significant wave height from synthetic
  aperture radar,''
\newblock {\em IEEE Transactions on Geoscience and Remote Sensing}, vol. 59,
  no. 3, pp. 1859--1867, 2020.

\bibitem{Colin-2021-IGARSS}
A.~Colin, C.~Peureux, R.~Husson, N.~Long{\'e}p{\'e}, R.~Rauzy, R.~Fablet,
  P.~Tandeo, S.~Saoudi, A.~Mouche, and G.~Dibarboure,
\newblock ``Segmentation of sentinel-1 sar images over the ocean, preliminary
  methods and assessments,''
\newblock in {\em Proceedings of IGARSS}. IEEE, 2021, pp. 4067--4070.

\bibitem{De-2021-JSTAR}
I.~De~Gelis, A.~Colin, and N.~Long{\'e}p{\'e},
\newblock ``Prediction of categorized sea ice concentration from sentinel-1 sar
  images based on a fully convolutional network,''
\newblock {\em IEEE Journal of Selected Topics in Applied Earth Observations
  and Remote Sensing}, 2021.

\bibitem{Wang-2019-GDJ}
C.~Wang, A.~Mouche, P.~Tandeo, J.~E. Stopa, N.~Longepe, G.~Erhard, R.~C.
  Foster, D.~Vandemark, and B.~Chapron,
\newblock ``A labelled ocean sar imagery dataset of ten geophysical phenomena
  from sentinel-1 wave mode,''
\newblock {\em Geoscience Data Journal}, vol. 6, no. 2, pp. 105--115, 2019.

\bibitem{Kim-2020-TIP}
W.~Kim, A.~Kanezaki, and M.~Tanaka,
\newblock ``Unsupervised learning of image segmentation based on differentiable
  feature clustering,''
\newblock {\em IEEE Transactions on Image Processing}, vol. 29, pp. 8055--8068,
  2020.

\bibitem{Wang-2020-GRSL}
Z.~Wang, X.~Fu, and K.~Xia,
\newblock ``Target classification for single-channel sar images based on
  transfer learning with subaperture decomposition,''
\newblock {\em IEEE Geoscience and Remote Sensing Letters}, pp. 1--5, 2020.

\bibitem{Brekke-2013-GRSL}
C.~Brekke, S.~N. Anfinsen, and Y.~Larsen,
\newblock ``Subband extraction strategies in ship detection with the
  subaperture cross-correlation magnitude,''
\newblock {\em IEEE Geoscience and Remote Sensing Letters}, vol. 10, no. 4, pp.
  786--790, 2013.

\bibitem{Singh-2010-IGARSS}
J.~Singh, M.~Soccorsi, and M.~Datcu,
\newblock ``Sar complex image analysis: A gauss markov and a multiple
  sub-aperture based target characterization,''
\newblock in {\em Proceedings of IGARSS}, 2010, pp. 1585--1588.

\bibitem{Focsa-2020-COMM}
A.~Focsa, M.~Datcu, S.A. Toma, A.~Anghel, and R.~Cacoveanu,
\newblock ``Opportunistic bistatic sar image classification using sub-aperture
  decomposition,''
\newblock in {\em Proceedings of COMM}, 2020, pp. 203--207.

\bibitem{Dhillon-2020-PAI}
A.~Dhillon and G.~K. Verma,
\newblock ``Convolutional neural network: a review of models, methodologies and
  applications to object detection,''
\newblock {\em Progress in Artificial Intelligence}, vol. 9, no. 2, pp.
  85--112, 2020.

\end{thebibliography}

\end{document}